\documentclass[11pt]{nokiatr}

\title{AnyJev Technical Report}
\subtitle{Typed Decisions from One Prefill of Any Open LLM}
\reportname{AnyJev}

\team{Nokia Applied Research}

\author{%
  Jiamu Zhang\textsuperscript{1}\quad Tianze Yang\textsuperscript{1}\quad
  Yucheng Shi\textsuperscript{2}\quad Evan Chen\textsuperscript{1} \\
  Zixiang Nie\textsuperscript{1}\quad Kelly Wan\textsuperscript{1}\quad Liangjie Hong\textsuperscript{1}\quad
  Ninghao Liu\textsuperscript{3}\quad Liang Wu\textsuperscript{1}%
}
\affiliations{%
  \textsuperscript{1}Nokia, Sunnyvale, CA, USA\qquad
  \textsuperscript{2}Tencent Hunyuan \\
  \textsuperscript{3}The Hong Kong Polytechnic University%
}

\links{%
  \githublink{https://github.com/nokia-applied-research/AnyJev}%
    {nokia-applied-research/AnyJev}
}

\date{}
\reportstatus{Early report on work in development}

\begin{document}
\begin{reportcover}
\maketitle
\begin{abstract}
A typed decision is a choice among a fixed set of options, returned as a probability rather than as
text. Systems that need typed decisions today use models trained for that purpose. This report
describes AnyJev, which reads a typed decision from one prefill of a pretrained instruction-tuned
language model. The readout restricts the next-token distribution at the answer position to the
option tokens. It has two defects: the model assigns higher probability to some labels whatever the
input, and to some positions in the option list. AnyJev corrects both with no gradient steps and no
parameter changes: it divides out a label prior estimated from unlabelled inputs, and it averages
log-probabilities over the $K$ cyclic rotations of the option list. On two 20-option tasks the
rotations lower the order-flip rate from 0.33 to 0.14 and from 0.33 to 0.18, and raise accuracy on
11 of 11 models on both. Reading every rotation requires $K$ prefills. A stopping rule selected against the full-rotation decision on unlabelled
states cuts that. Selecting the threshold on one unlabelled split and bounding its disagreement on a
second, it reads 10.6 rotations of 18 at a verified 0.008 bound on two of four cells; selected and
bounded on one split, as our serving run did, it reads 7.3 and serves 2.2 times as many decisions
per second on vLLM. The code is open source.
\end{abstract}

\end{reportcover}

\section{Introduction}

An agent system has to make many small decisions that are not text: route this request to one of
twenty skills, decide whether this input is an injection attempt, score this answer from one to
five. A language model can be asked for these in prose, but the caller then has to parse the reply,
and gets no probability it can threshold. A \emph{typed decision} fixes the interface instead. The
caller supplies a state and a question with a fixed set of options, and the system returns a
distribution over those options. The interface was made popular by a hosted service, and open
implementations now include both trained decision models and frozen-model readouts
(Section~\ref{sec:related-work}).

\begin{figure}[!t]
  \centering
  \includegraphics[width=0.93\textwidth]{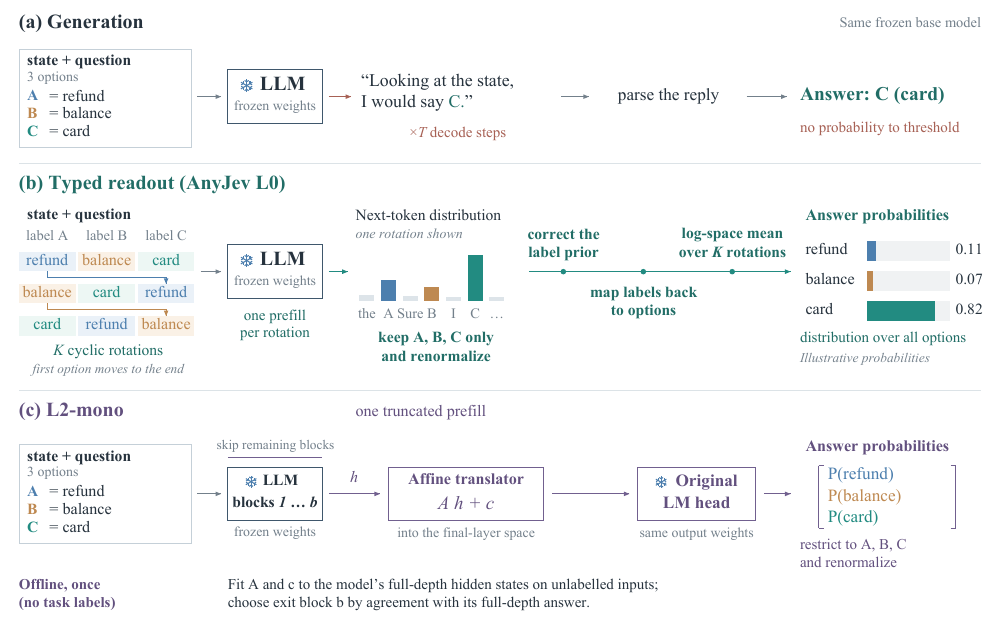}
  \caption{The same base model, three decision paths. (a) Asking for the answer in text costs one forward pass
  per generated token and returns prose the caller has to parse, with no probability attached.
  (b) AnyJev L0 runs one prefill per cyclic rotation. It restricts the next-token distribution
  to the option-label tokens and renormalises, corrects the label prior, maps each position back
  to its option, and averages in log space across the $K$ rotations before renormalising.
  A, B and C are position labels whose option assignments change across rotations; the final
  distribution is indexed by option. Probabilities are illustrative, and one rotation's token
  distribution is shown. L0 debiases the readout; confidence calibration is a separate L1 step.
  (c) L2-mono truncates the forward at a selected block, translates the hidden state into the
  final-layer space with an affine map, and uses the original output head for the option readout.
  The translator is fitted once on unlabelled inputs to the model's full-depth hidden states;
  the exit depth is selected by agreement with its full-depth answers, without task labels.
  The L2-mono output is symbolic. All three paths keep the base-model weights fixed.}
  \label{fig:teaser}
\end{figure}

This report takes a different position (Figure~\ref{fig:teaser}). The capability is already
present in the pretrained
instruction-tuned models we tested, and what needs correcting is the readout rather than the model.
AnyJev builds one prompt from the state and the question, runs a single prefill, and reads
the decision from the next-token distribution at the answer position, restricted to the tokens that
name the options. No parameter changes, no generated tokens, and no second model.

Read that way, the distribution is usable and wrong in two specific ways. The model assigns higher
probability to some labels whatever the state says, and it assigns higher probability to some
positions in the option list. The second defect is the one a caller notices. On a 20-option task,
reversing the option list changes the answer on 33\% of items, averaged over eleven models. Both
defects are properties of the readout rather than of the model's knowledge, Section~\ref{sec:correction} corrects both with no gradient
steps, from the model's own outputs on unlabelled inputs.

Accuracy varies with the base model and changes little under the corrections. On the public
subset of a typed-decision benchmark the eight models we ran span 0.549 to 0.798, and the debiasing
does not change accuracy there by a measurable amount. It supplies a large reduction in order dependence and, with a
small labelled set, calibrated confidence. Section~\ref{sec:results} separates the two and reports each with the interval it was
measured to.

\section{Related Work}
\label{sec:related-work}

\subsection{Typed decision models and open implementations}
Jev popularizes probabilistic \texttt{Choice}, \texttt{Score}, and \texttt{Noul}
interfaces~\cite{almeida2026jev}, building on the broader connection between language-model
predictions and classification~\cite{schick2021pet,gao2021lmbff}. Frozen-model implementations
include Yinsongxu's LLM2Jev, which scores candidates independently, tic-top's llm2jev, which
reads labels from a joint option prompt, and SemIf, which supports shared-state
execution~\cite{llm2jev,llm2jev-joint,semif}.
NanoJev, Kev, Laya, and related releases explore learned scorers and local decision
services~\cite{nanojev,kev,nimble,laya,decider,jevlike,jevos}.
Visual Jev extends shared-context decisions to images, while NumericJev obtains numerical
outputs through multi-round interval refinement~\cite{yu2026visualjev,ye2026numericjev}.

\subsection{Probability readouts and calibration}
Label-probability readouts inherit surface-form and option-order
biases~\cite{holtzman2021surface,zheng2024robust}. AnyJev combines prior correction related
to content-free and batch calibration with temperature
scaling~\cite{zhao2021calibrate,zhou2024batch,guo2017calibration}, while measuring the prefill
cost of averaging layouts. Recent Jev studies expose sensitivity to option-name semantics
and inconsistency across logically linked questions~\cite{sun2026typesafe,li2026coherence}.
These distinguish schema validity from semantic correctness and probability quality;
AnyJev focuses on label and position bias and calibrated confidence on frozen backbones.

\subsection{Decision components in agent systems}
Model cascades and Jev-based judging use confidence to accept or escalate decisions, although
correlated errors can limit fallback gains~\cite{chen2024frugalgpt,wu2026reflex,li2026jevjudge,rao2026rubricjudges}.
Open applications use typed decisions for browser control, history pruning, and code
retrieval~\cite{jev-ultrafast,fast-jev-compaction,jevgrep}.
Such components complement Nokia's KV-cache compression, context management, and MoE
paging work~\cite{zhang2026spectra,min2026trace,zhang2026wisp}.
AnyJev could also support production agents in autonomous networks, motivated by AI-native
6G visions and Jev-based edge orchestration~\cite{wu2026autonomousnetworks,li2026sixgorchestration}.
Self-Evolving Code-with-Image Reasoning and Field-Aware Agent Skill Retrieval suggest roles
in solver invocation and skill selection~\cite{yang2026codewithimage,goulart2026fieldaware};
recommendation and time-series systems provide further domain
contexts~\cite{wang2026pagellm,zheng2026sapo,chen2026timeseries}. These proposed combinations
require application-level evaluation.

\section{The readout}
\label{sec:readout}

\paragraph{Prompt.} A decision has a state $s$ and a question with $K$ options $o_1,\dots,o_K$.
AnyJev gives each \emph{position} in the printed list a single-token label $\ell_1,\dots,\ell_K$:
letters for a choice, \texttt{Yes} and \texttt{No} for a binary question, digits for an ordinal
score. A layout $\pi$ maps positions to options, so position $j$ shows option $\pi(j)$. The prompt
$x(s,\pi)$ places a fixed instruction and the state, then the question, then one line
\texttt{$\ell_j$. $o_{\pi(j)}$} per position, then a line telling the model to answer with the label
only. The state precedes the options, so a backend can reuse the prefix across layouts of the same
question.

\paragraph{Decision.} We run one prefill of $x(s,\pi)$ and read the next-token distribution $p$ at
the last position. Only the $K$ label tokens can be answers, so we restrict and renormalise:
\begin{equation}
  m(s,\pi) \;=\; \sum_{j=1}^{K} p(\ell_j), \qquad
  \tilde p_j \;=\; \frac{p(\ell_j)}{m(s,\pi)} .
  \label{eq:restrict}
\end{equation}
The decision is $\tilde p$ mapped from positions back to options. The model generates nothing, so
one decision costs one prefill. Scoring single-token labels rather than the option strings also
removes surface-form competition, where a correct option receives lower probability because it is
spelled in more tokens or rarer ones~\cite{holtzman2021surface}.

\paragraph{Answer mass.} The normaliser $m$ is the share of the model's own probability that reaches
the option tokens, and it reports whether the last prompt token is the position the model would
answer at. A chat template that puts something else there still yields an argmax over the label
tokens, and that argmax can still be correct often enough to resemble a working readout. We report
$m$ before any accuracy and exclude a row whose $m$ is low rather than average it in. One template
fault was found this way: the harmony format used by gpt-oss ends the prompt at the assistant
header, before the channel marker, which gave $m = 0.000$ while the answers still looked plausible.

Every benchmark row here has median $m \geq 0.995$ (Table~\ref{tab:jevbench}). On the 20-option
sweep three rows fall below 0.99, all on the binary \texttt{injection} task: 0.408 on SmolLM2-1.7B,
0.710 on OLMo-2-7B and 0.976 on Granite-3.3-8B. We exclude those three from every statistic for that
task. Appendix~\ref{app:breadth} prints $m$ for every row.

\paragraph{Two defects.} The model assigns higher probability to some labels whatever the state
says, higher to \texttt{Yes} than to \texttt{No} and to earlier letters than to later ones. It also
assigns higher probability to some positions in the printed list, so the decision depends on the
order the caller wrote the options in. Averaged over eleven models, reversing the option list
changes the argmax on 32.8\% of \texttt{banking20} items and on 33.4\% of \texttt{newsgroups} items.

\section{Correcting the two defects}
\label{sec:correction}

\subsection{The label prior}

The label prior is a distribution $\hat q$ over positions that the model applies regardless of the
state. We estimate it from unlabelled inputs and divide it out:
\begin{equation}
  p^{\mathrm{dbs}} \;\propto\; \tilde p \,/\, \hat q^{\,\alpha},
  \label{eq:prior}
\end{equation}
renormalised over the $K$ options, with $\alpha \in [0,1]$ setting how much of the prior we remove.
We take $\hat q$ to be the running mean of $\tilde p$ over the inputs the system has already scored,
which is batch calibration~\cite{zhou2024batch}, held per question and per layout. The estimate uses
the model's own outputs and no gold label. A second estimator computes $\hat q$ from content-free
probes instead~\cite{zhao2021calibrate}; AnyJev offers it and we do not use it here.

We set $\alpha = 0.75$, taken from a separate sweep over 230 (model, question) pairs scored against gold labels, so $\alpha$ is the second place a label enters this system, alongside
the temperature of Section~\ref{sec:confidence}. Eq.~\eqref{eq:prior} also assumes the label
marginal of the inputs is not extreme, and it lowers accuracy where that assumption fails.

\subsection{Position bias}

Position bias needs no estimator, because the layout is ours to choose. We read the same decision
under the $K$ cyclic rotations $\pi_t(j) = (j + t) \bmod K$, so that every option occupies every
position once, and we average in log space~\cite{zheng2024robust}:
\begin{equation}
  \bar z_i \;=\; \frac{1}{K}\sum_{t=0}^{K-1} \log q_t(i), \qquad
  p^{L0} \;=\; \operatorname{softmax}(\bar z),
  \label{eq:rot}
\end{equation}
where $q_t(i)$ is the restricted probability of Eq.~\eqref{eq:restrict} for option $i$ under
rotation $t$, before the label prior.

The log-space average makes the correction exact under a stated model of the bias. Suppose the logit
for option $i$ at position $j$ separates as $c_i + b_j$. Then
$\log q_t(i) = c_i + b_{j(t,i)} - \log Z_t$, where $Z_t$ normalises rotation $t$. Summing over all
$K$ rotations sends option $i$ through every position once, so $\sum_t b_{j(t,i)} = \sum_j b_j$ for
every $i$, and the two remaining terms do not depend on $i$. Hence
$\operatorname{softmax}(\bar z) = \operatorname{softmax}(c)$ exactly. Averaging probabilities instead
of their logarithms does not have this property.

Two qualifications follow. The label prior is held per layout, so applying it before the average
contributes a further term that cancels only when $\hat q$ depends on position alone; AnyJev
therefore averages the pre-prior quantity. And the additive model is an approximation. Under it
reversing the option list would change nothing, and reversal still changes the answer on 0.138 of
\texttt{banking20} items and 0.183 of \texttt{newsgroups} items after the average
(Table~\ref{tab:breadth}).

\paragraph{Canonical order.} Reading every rotation gives each option each position, but it does not
fix which options are adjacent, and the options attend to one another inside the prompt. Sorting the
option list by its text before rotating it removes that remaining dependence: the prompt set becomes
a function of the option \emph{set}, so any listing of the same options produces the same prompts
and the same probabilities, at any rotation budget, which follows from the construction. The runs
reported here rotate the caller's listing, so we do not measure it.

\subsection{Confidence}
\label{sec:confidence}

Neither correction makes the model's confidence match its accuracy, and we did not measure a
label-free procedure that detects this. One temperature, fitted on 100 to 500 labelled examples of
the same question, is the correction AnyJev applies for it~\cite{guo2017calibration}. A single
positive temperature rescales $\bar z$ and leaves the argmax unchanged. The level we report as
\emph{+temperature} also re-estimates $\hat q$ from the calibration set alone, so its accuracy
differs from the debiased level by up to 0.013; we report it for calibration and read accuracy off
the debiased level.

\section{Reading fewer rotations}
\label{sec:budget}

Eq.~\eqref{eq:rot} requires $K$ prefills per decision. On the 18-option routing question used for
timing, that is 18 prefills where the naive readout uses one. Two changes can reduce it: make each
rotation cheaper, or read fewer of them.

Section~\ref{sec:negative} reports what the first achieved. The $K$ prompts share every token up to
the option list, so an engine that caches the shared prefix already recovers most of the cost, and
packing all $K$ rotations into one forward then runs 1.20 times as fast as that path on a short
state. This section reads fewer rotations instead.

\paragraph{Reading order.} Consecutive rotations are nearly the same layout, because $\pi_{t+1}$
moves every option one position from $\pi_t$. AnyJev therefore reads rotations in the order
$0,\,K/2,\,K/4,\,3K/4,\,K/8,\dots$, the van der Corput sequence scaled to $K$, so that a short
prefix places each option in well separated positions.

\paragraph{When to stop.} After reading a subset of rotations we hold a running average $\bar z$.
The obvious statistic is the gap between its top two probabilities, and that gap saturates at 1.
Once the average is peaked it no longer separates a settled decision from one settled by a wide
margin, which is the regime the rule acts in. We use the same gap in log space,
\begin{equation}
  \mu(\bar z) \;=\; \bar z_{(1)} - \bar z_{(2)},
  \label{eq:margin}
\end{equation}
which is unbounded and unchanged by renormalisation of $\bar z$. The rule stops at the first
rotation where $\mu \geq \tau$, after a minimum of two.

\paragraph{Choosing the threshold without labels.} The guarantee we can state without labels is not
about accuracy. We state it against our own full-rotation decision instead: \emph{the budget returns
the decision that reading every rotation would have returned, on at least $1-\varepsilon$ of
inputs}. That target needs only the model's own output on unlabelled states.

Given $n$ unlabelled states we read every rotation of each once. For a candidate $\tau$ let
$d(\tau)$ count the states whose stopped decision differs from the full-rotation decision, and take
the smallest $\tau$ on a grid for which
\begin{equation}
  U\bigl(d(\tau),\, n,\, \delta\bigr) \;\leq\; \varepsilon ,
  \label{eq:cert}
\end{equation}
where $U$ is the Clopper--Pearson upper confidence bound~\cite{clopper1934confidence} at confidence
$1-\delta$, with $\delta = 0.05$. If no $\tau$ satisfies Eq.~\eqref{eq:cert}, the question reads
every rotation.

Eq.~\eqref{eq:cert} is evaluated at every candidate on the grid, so a bound taken from the states
that chose $\tau$ applies to each candidate separately and not to the selected one. We therefore
select and verify on disjoint states: $\tau$ comes from one third of an unlabelled batch, and the
bound the target must clear is computed on the other two thirds, for that single threshold. The
arithmetic sets the split. At $\delta = 0.05$ a verification split of $n$ states with one
disagreement gives a bound near $4.7/n$, so a 1\% target needs roughly 500 states, and 300 cannot
certify 1\% unless no disagreement occurs.

On four cells the procedure certifies two. Qwen2.5-7B reaches a verified bound of 0.0079 on both
tasks, at 10.61 of 18 rotations and 9.01 of 20, which is 1.70 and 2.22 times fewer
(Table~\ref{tab:cert-split}). Qwen3-8B on newsgroups misses at 0.0105, and on
\texttt{massive\_route} no threshold on the grid clears the target on the selection split. Selecting
and verifying on the same states, as the runs in Section~\ref{sec:results} did, returns a smaller
threshold and a larger saving, and its bound does not cover the selection.

Table~\ref{tab:certificate} reports which statistic satisfies Eq.~\eqref{eq:cert} at
$\varepsilon = 0.01$. The log-odds margin certifies all four cells; the probability gap certifies
two, and on the other two no threshold clears the target at any budget.

\paragraph{The cost of a round.} The budget requests rotations from the engine in rounds, and every
request in a batch must finish a round before the next starts. Reading one rotation per round issues
the fewest forwards and the most rounds. Section~\ref{sec:results} measures both, and the two
engines rank them differently.

\begingroup
\emergencystretch=1em
\section{Experimental setup}
\label{sec:setup}

\paragraph{Models.} For the 20-option, JevBench and serving experiments, we ran thirteen
instruction-tuned models from eight organisations at their
released weights, in \texttt{bfloat16}: Qwen3 at 1.7B, 4B, 8B, 32B and
30B-A3B~\cite{yang2025qwen3}, Qwen2.5-7B~\cite{qwen2024qwen25},
Llama-3.2-3B~\cite{grattafiori2024llama3}, Mistral-7B-v0.3~\cite{jiang2023mistral},
Granite-3.3-8B~\cite{granite2024granite}, Phi-4-mini~\cite{abouelenin2025phi4mini},
SmolLM2-1.7B~\cite{benallal2025smollm2}, OLMo-2-7B~\cite{olmo2024olmo2} and
gpt-oss-20b~\cite{openai2025gptoss} at its low reasoning setting. The 20-option sweep covers eleven
of them from six organisations, the benchmark run eight from five. We fine-tuned, quantised and
modified nothing.

\paragraph{Tasks.} The 20-option sweep uses three tasks: \texttt{banking20}, twenty banking intents
from BANKING77~\cite{casanueva2020efficient}; \texttt{newsgroups}, the twenty classes of 20
Newsgroups~\cite{lang1995newsweeder}; and \texttt{injection}, a binary question over prompt
injections. Each cell uses 300 held-out items. Timing uses an 18-option utterance routing question
built from MASSIVE~\cite{fitzgerald2023massive}.

\paragraph{Benchmark.} We also ran the public subset of JevBench~\cite{jevbench2026}, a public set of typed
decisions for this class of system. It publishes 231 items in three files. Its overall score averages three
difficulty tiers that partition a larger set than the public files, so we report the public subset
and do not place it against leaderboard entries. We score 213 of the 231. The other 18 state an
expected value that is not one of the listed options: twelve are ordinal and six are otherwise
irregular, and we skip them rather than guess a mapping.

\paragraph{What is fitted on what.} The rotation average uses no labels. Two quantities are fitted
against gold labels and we name both: the prior strength $\alpha$, selected once on a separate sweep
(Section~\ref{sec:correction}), and the temperature, fitted out of fold over items, five folds taken
over the pooled items rather than within a difficulty tier. A temperature fitted on one tier alone
would misfit the others. The stopping threshold of Section~\ref{sec:budget} is selected on unlabelled
states of the same question.

\paragraph{Settings and hardware.} The main tables read every rotation, combine in log space, and
rotate the caller's listing rather than a canonical one. The 20-option sweep applies the label prior
of Eq.~\eqref{eq:prior}; the benchmark run does not, so its debiased column isolates the rotations.
These experiments used one H100 NVL, PyTorch 2.5.1 with CUDA 12.4 and Transformers
4.55.4~\cite{wolf2020transformers}. The serving comparison additionally used vLLM
0.7.0~\cite{kwon2023pagedattention} on the same host, measured serially on an otherwise idle
machine, because an earlier pair of runs that shared the host returned a different ratio.

\paragraph{Additional full-split evaluation.} We separately evaluate the default BEV Decision Mix
test split on two AnyJev base models and four published baseline interfaces. Its readouts,
calibration protocol and software settings are specified in Appendix~\ref{app:bev-full}; the
temperature fit and hardware configuration above do not describe that separate evaluation.

\endgroup
\section{Results}
\label{sec:results}

\subsection{Order dependence falls on every model tested}

\begin{figure}[t]
  \centering
  \includegraphics[width=0.88\textwidth]{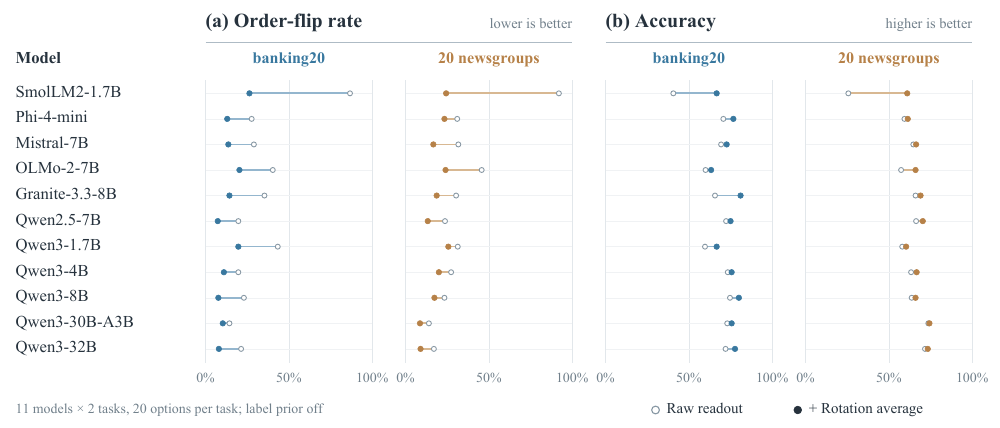}
  \caption{Rotation averaging on two 20-option tasks. Each row is one model, with separate
  subpanels for \texttt{banking20} and 20 newsgroups under each metric. Hollow points show the
  raw readout; filled points show the rotation average of Eq.~\eqref{eq:rot}.
  Left: the order-flip rate, the share of items whose answer changes when the option list is
  reversed. Right: accuracy. Axes are shown as percentages. All 22 model--task pairs have a
  lower order-flip rate and higher accuracy after averaging. The label prior is switched off,
  so the changes are due to rotations alone.}
  \label{fig:breadth}
\end{figure}

\begin{table}[t]
  \centering
  \small
\begin{tabular}{l ccc}
\toprule
 & banking20 & 20 newsgroups & injection \\
\midrule
 & $K=20$ & $K=20$ & $K=2$ \\
\midrule
\textit{Accuracy} &  &  &  \\
\quad raw readout & 0.665 & 0.609 & 0.720 \\
\quad + rotations & 0.737 & 0.668 & 0.733 \\
\quad + rotations + label prior & 0.752 & 0.675 & 0.762 \\
\textit{Order-flip rate} &  &  &  \\
\quad raw readout & 0.328 & 0.334 & 0.180 \\
\quad + rotations & 0.138 & 0.183 & 0.000 \\
\textit{Calibration error} &  &  &  \\
\quad raw readout & 0.251 & 0.276 & 0.202 \\
\quad + rotations + label prior & 0.223 & 0.285 & 0.170 \\
\quad + temperature & 0.088 & 0.099 & 0.095 \\
\midrule
Models passing the answer-mass gate & 11 of 11 & 11 of 11 & 8 of 11 \\
Rotations raise accuracy on & 11/11 & 11/11 & 3/7 \\
Rotations lower the flip rate on & 11/11 & 11/11 & 8/8 \\
\bottomrule
\end{tabular}

  \caption{Means over the models that pass the answer-mass gate, 300 items per cell. The lower block
  counts models rather than items. Under a coin flip, 11 of 11 occurs with probability
  $4.9\times10^{-4}$ (exact sign test, one-sided). The injection accuracy row is 3 of 7 after
  dropping one tie, at $p=0.77$.}
  \label{tab:breadth}
\end{table}

The rotations lower the order-flip rate on 11 of 11 models on both 20-option tasks, from 0.328 to
0.138 on \texttt{banking20} and from 0.334 to 0.183 on \texttt{newsgroups}
(Table~\ref{tab:breadth}, Figure~\ref{fig:breadth}). They raise accuracy on 11 of 11 models on both,
by 7.2 and 5.9 points on the means. An exact sign test over the eleven models puts each of those
counts at $p = 4.9\times10^{-4}$.

The size of the effect depends on the number of options. On the binary \texttt{injection} task the
rotations remove order dependence completely, from 0.180 to 0.000 on the eight rows that pass the
answer-mass gate, and they do not raise accuracy: they win on 3 of the 7 rows that are not ties, at
$p = 0.77$. With two options there are two positions, and little position bias to average away.

Calibration moves under the temperature rather than under the rotations. The rotations and the prior
take the calibration error from 0.251 to 0.223 on \texttt{banking20}, and the temperature takes it
to 0.088.

\subsection{The benchmark}

\begin{figure}[t]
  \centering
  \includegraphics[width=0.88\textwidth]{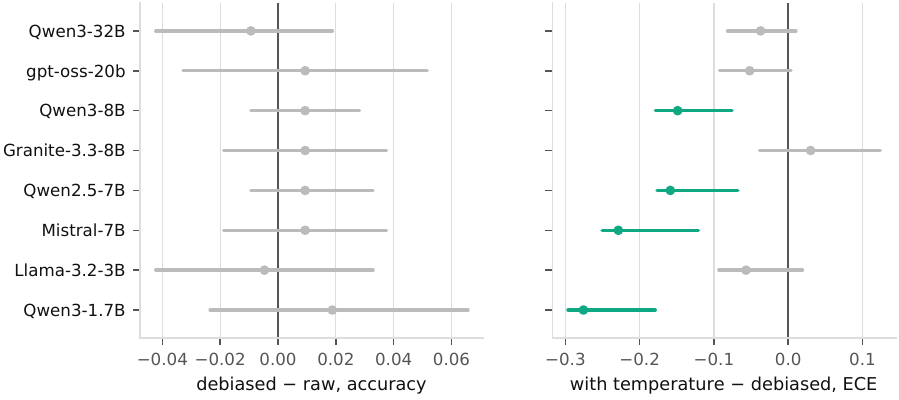}
  \caption{Differences on the JevBench public subset with 95\% bootstrap intervals over 2\,000
  resamples, paired over items, one row per model. Left: debiased minus raw accuracy; every interval
  contains zero. Right: the change in calibration error from adding the temperature; four of the
  eight intervals exclude zero, and all four are improvements. Coloured intervals exclude zero.}
  \label{fig:jevbench}
\end{figure}

\begin{table}[t]
  \centering
  \small
\begin{tabular}{l cc c ccc}
\toprule
Model & Answer mass & Raw & Debiased & ECE raw & ECE debiased & ECE + temp. \\
\midrule
Qwen3-32B & 1.000 & 0.798 & 0.789 & 0.140 & 0.128 & 0.091 \\
Qwen3-8B & 1.000 & 0.704 & 0.714 & 0.278 & 0.263 & 0.114 \\
gpt-oss-20b & 0.999 & 0.704 & 0.714 & 0.140 & 0.154 & 0.102 \\
Granite-3.3-8B & 0.999 & 0.690 & 0.700 & 0.280 & 0.270 & 0.301 \\
Qwen2.5-7B & 0.996 & 0.676 & 0.685 & 0.259 & 0.243 & 0.085 \\
Mistral-7B & 0.999 & 0.615 & 0.624 & 0.322 & 0.333 & 0.104 \\
Llama-3.2-3B & 0.999 & 0.563 & 0.559 & 0.174 & 0.163 & 0.107 \\
Qwen3-1.7B & 1.000 & 0.549 & 0.568 & 0.395 & 0.380 & 0.105 \\
\bottomrule
\end{tabular}

  \caption{JevBench public subset, 213 items scored per model. Answer mass is the normaliser of
  Eq.~\eqref{eq:restrict}. ECE is expected calibration error over 15 equal-mass bins. Intervals for
  the differences are in Figure~\ref{fig:jevbench} and Table~\ref{tab:jevbench-ci}.}
  \label{tab:jevbench}

\end{table}

Accuracy on this subset spans 0.549 for Qwen3-1.7B to 0.798 for Qwen3-32B
(Table~\ref{tab:jevbench}). The spread belongs to the models. The debiasing does not change
accuracy here by a measurable amount: no per-model interval excludes zero, and pooling all
$8\times213 = 1704$ decisions gives a difference of $+0.006$ with a 95\% interval of
$[-0.004, +0.018]$.

Section~\ref{sec:correction} predicts that the effect grows with the number of options, and this
benchmark has few. Its items
have 2 to 6 options, 3.6 on average over the 213 scored.
Table~\ref{tab:by-k} splits the pooled decisions by option count. The $K=6$ interval excludes zero, but it is one of five intervals. Were they
independent, the chance that at least one of five 95\% intervals excludes zero under the null would
be 0.23; they share models and the resampling is clustered, so that figure is indicative rather
than exact. Testing the trend once, by regressing the per-decision difference on $K$ and
bootstrapping the slope clustered by item, gives $+0.0059$ accuracy per additional option with an
interval of $[-0.0012, +0.0132]$. The sign matches the prediction and the sample does not resolve
it. We therefore report no accuracy effect on this benchmark.

The temperature lowers calibration error on seven of the eight models and raises it on one. Four of
the eight intervals exclude zero, all of them improvements: Qwen3-8B, Qwen2.5-7B, Mistral-7B and
Qwen3-1.7B, between $-0.149$ and $-0.276$. On Granite-3.3-8B the error rises from 0.270 to 0.301,
with an interval that contains zero. One global temperature cannot correct a model whose confidence
is wrong in different directions on different items.

\subsection{Full BEV Decision Mix evaluation}

On the full test split of the default BEV Decision Mix configuration~\cite{bevdecision2026},
all systems are scored on the same 46\,320 decisions (Appendix~\ref{app:bev-full},
Table~\ref{tab:bev-full}). With the label prior off, rotation averaging reaches 68.01\% on
Qwen3.5-9B and 67.59\% on Qwen3-32B, compared with 67.85\% and 67.03\% for their raw readouts.
Bespoke-Nimble-9B reaches 70.57\%; its 2.56-point lead over the 9B rotation average is an observed
system difference, not an isolated estimate of the effect of fine-tuning. Adding the batch prior
at $\alpha=0.75$ lowers accuracy to 65.01\% and 65.55\%, respectively, although it improves the
9B model's ECE from 0.0490 to 0.0418. Thus the prior's accuracy benefit does not transfer to this
evaluation. The evaluated affine-map L2-mono variant reads block 29 of 32 and scores 67.24\%,
0.61 points below the same model's full-depth, single-forward readout; latency was not measured.

\subsection{Serving cost}

\begin{table}[t]
  \centering
  \small
\begin{tabular}{l l ccc cc}
\toprule
Engine & Readout & Decisions/s & Speed-up & Rotations read & Accuracy & Agreement \\
\midrule
vLLM & all 18 rotations & 16.7 & 1.00 & 18.0 & 0.697 & 1.000 \\
 & budget, wave 1 & 22.1 & 1.32 & 7.2 & 0.703 & 0.987 \\
 & budget, wave 2 & 37.2 & 2.22 & 7.3 & 0.703 & 0.987 \\
 & budget, wave 4 & 31.8 & 1.90 & 9.3 & 0.703 & 0.987 \\
 & budget, wave 6 & 29.2 & 1.74 & 10.9 & 0.700 & 0.987 \\
\midrule
Transformers & all 18 rotations & 7.0 & 1.00 & 18.0 & 0.697 & 1.000 \\
 & budget, wave 1 & 16.5 & 2.34 & 7.2 & 0.703 & 0.987 \\
 & budget, wave 2 & 16.5 & 2.34 & 7.4 & 0.703 & 0.987 \\
 & budget, wave 4 & 13.1 & 1.87 & 9.2 & 0.703 & 0.987 \\
 & budget, wave 6 & 11.4 & 1.62 & 10.8 & 0.700 & 0.990 \\
\bottomrule
\end{tabular}

  \caption{300 decisions of an 18-option question on one H100. The wave is how many rotations a
  decision requests per engine call. Accuracy spans 0.697 to 0.703 across the rows, and agreement is
  measured against the full-rotation decision. Speed-up is against the all-rotations row of the same
  engine.}
  \label{tab:throughput}

\end{table}

On the 18-option routing question, Eq.~\eqref{eq:cert} returns a threshold of 6.0 from 600
unlabelled states, with one disagreement and a Clopper--Pearson upper bound of 0.0079 against a 1\%
target. Reading two rotations per engine call, the budget then reads 7.3 of the 18 and returned the
full-rotation decision on 98.7\% of held-out decisions (Table~\ref{tab:throughput}). That is a
disagreement rate of 0.013, or 4 decisions of 300. This run selected its threshold and took its
bound from the same states, so the bound does not cover the selection, and the rate is the one to
read. Section~\ref{sec:budget} reports the same cell under selection and verification on disjoint
states: the threshold rises from 6.0 to 10.25, the rotations read from 7.3 to 10.6, and the bound
becomes one a deployment can rely on.
We quote 6.0 here because it is the threshold the serving run used.

The round size matters as much as the rotation count, and the two engines rank it differently. On
vLLM, reading one rotation per round issues the fewest rotations, 7.2 of 18, and reaches 22.1
decisions per second. Reading two per round issues 7.3 and reaches 37.2, which is 2.22 times the
all-rotations rate. At four per round the rate falls to 31.8. On Transformers the first two rounds
sizes are level at 16.5 decisions per second, 2.34 times the all-rotations rate, and larger rounds
are slower.

We also measured whether the decision can be read from a truncated forward, by fitting a label-free
affine map from an intermediate block into the final basis~\cite{belrose2023tuned}. At a 0.98
agreement target six of the eight models save no blocks. Appendix~\ref{app:exit} reports the depths.

\section{What did not work}
\label{sec:negative}

Each mechanism below does what its design says, to the precision quoted, and a measurement rather
than a fault stopped each one.

\paragraph{A question-agnostic head.} A closed-form head fitted on one question's gold labels
against the model's hidden states scores 0.13 above the debiased readout at the same depth, and it
is fitted per question. Leaving out one of twenty questions at a time and fitting on the other
nineteen, four extraction variants, five depths and four functional forms all score below the raw
readout on the held-out question. The best reaches 0.580, against 0.635 for the debiased readout and
0.628 for the raw one, and its flip rate
under reversal is 0.17 to 0.30, against 0.07 for the raw readout on the same questions. The signal is
present per question; a shared linear rule that reaches an unseen question is not.

\paragraph{Reusing the question block's cache.} The question and its options are identical across
requests to a given endpoint, but they follow the state, so no prefix cache reaches them. On a short
state they are most of the prompt: 159 fixed tokens of which 42 precede the state, against a median
routing utterance of 7 tokens. Splicing the cached block into a new request and rotating its
positions is exact to $7.0\times10^{-5}$ in fp32 across all 28 layers. The reused block then
reproduces the previous state's decision in the new request.

\paragraph{Making the options a set.} The $K$ rotations exist because the options are written in
sequence, so option 5 attends to options 1 through 4 and option 1 to none of them. A mask in which
every option span sees the prefix and itself, each starting at the same position index, makes the
options exchangeable by construction: shuffling the spans moves the output by $3.7\times10^{-6}$
under it, against $2.2\times10^{-3}$ under the causal mask. Accuracy then falls to chance. On
Qwen2.5-7B and the 18-option routing question it scores 0.042 against 0.633 for the rotation
average, where chance is 0.056. Masking the options from each other while leaving their positions
alone already scores 0.092, so the loss comes from the blinding rather than the shared positions.
The model uses the other options when it selects among them.

\paragraph{Packing the rotations into one forward.} Laying the $K$ option blocks end to end, each at
the same starting position index and masked to see only the prefix and itself, reproduces $K$
independent prefills to $2.2\times10^{-5}$ and returns an identical decision. It runs 5.9 times as
fast as $K$ independent prefills, and 1.20 times as fast as the shared-prefix path an engine already
provides on a 10-token state, 1.11 times on a 200-token state. This is why
Section~\ref{sec:budget} reduces the number of rotations instead of their unit cost.

\section{Limitations}
\label{sec:limits}

\begin{itemize}[leftmargin=1.4em,itemsep=2pt,topsep=3pt]
  \item \textbf{One prefill sets a ceiling.} The readout generates nothing, so it cannot reason. On
        the hardest tier of the JevBench subset the eight models span 0.343 to 0.590.
  \item \textbf{JevBench has few options.} Its scored items average 3.6 options, the range where the rotation average has least to remove. We do not have a public many-option
        benchmark, and one would test the mechanism where it operates.
  \item \textbf{Eighteen JevBench items are unscored.} Twelve of the 231 public items are ordinal, where the
        expected value names a level rather than one of the listed options.
  \item \textbf{The 20-option results have no intervals.} Those runs record aggregates rather than
        per-item outcomes, so the strongest statement available from them is the sign test over
        eleven models.
  \item \textbf{The intervals are over items, not over runs.} They state what a different sample of
        items would have given. They do not cover seed, host, or the effect of batch composition on
        \texttt{bfloat16} logits.
  \item \textbf{The threshold is selected and certified on the same sample.} The bound therefore
        covers each candidate on the grid and not the selected one. On the serving run the held-out
        rate was 0.013 on 300 decisions, whose 95\% interval $[0.004, 0.034]$ contains the 0.01
        target, so that run neither confirms nor refutes it. Selecting on one unlabelled split and
        measuring on a second would settle it; we did not do that here.
  \item \textbf{The efficiency and accuracy measurements use different tasks.} The JevBench subset's $K$
        is 2 to 6, so a rotation budget has little to save there.
  \item \textbf{Hidden-state readouts need a local backend.} The affine map of
        Appendix~\ref{app:exit} and the per-question head need hidden states, which a hosted API
        does not expose. The readout of Section~\ref{sec:correction} needs only next-token
        log-probabilities.
\end{itemize}

\section{Conclusion}

The decision capability measured here is the base model's. What AnyJev supplies is a readout with
the label prior divided out and the position term averaged away, at one prefill per rotation and no
parameter changes. On both 20-option tasks that readout lowered the order-flip rate on every model
we tested, from 0.33 to 0.138 and 0.183, and a stopping rule selected against its own full-rotation
decision reduced the number of rotations read. AnyJev is open source.

\paragraph{Status.} This is an early report on work that is still in development. It covers the
readout, the corrections and their serving cost; the labelled head is not evaluated. The evidence
includes the 213-item JevBench public subset, three tasks of our own, and 46\,320 decisions from
the default BEV Decision Mix test split, including comparisons with published decision systems.
On that full split, the best AnyJev accuracy uses rotations without the label prior; the prior's
benefit is therefore task-dependent. These are early measurements, not a settled ranking of
decision systems.

\bibliographystyle{plain}
\bibliography{refs}

@misc{abouelenin2025phi4mini,
  author       = {Abdelrahman Abouelenin and Atabak Ashfaq and Adam Atkinson and others},
  title        = {Phi-4-Mini Technical Report: Compact yet Powerful Multimodal Language Models via Mixture-of-{LoRAs}},
  year         = {2025},
  eprint       = {2503.01743},
  archivePrefix = {arXiv},
  howpublished = {arXiv preprint arXiv:2503.01743}
}

@misc{belrose2023tuned,
  author       = {Nora Belrose and Igor Ostrovsky and Lev McKinney and Zach Furman and
                  Logan Smith and Danny Halawi and Stella Biderman and Jacob Steinhardt},
  title        = {Eliciting Latent Predictions from Transformers with the Tuned Lens},
  year         = {2023},
  eprint       = {2303.08112},
  archivePrefix = {arXiv},
  howpublished = {arXiv preprint arXiv:2303.08112}
}

@misc{benallal2025smollm2,
  author       = {Ben Allal, Loubna and Lozhkov, Anton and Bakouch, Elie and others},
  title        = {{SmolLM2}: When Smol Goes Big -- Data-Centric Training of a Small Language Model},
  year         = {2025},
  eprint       = {2502.02737},
  archivePrefix = {arXiv},
  howpublished = {arXiv preprint arXiv:2502.02737}
}

@inproceedings{casanueva2020efficient,
  author       = {I{\~n}igo Casanueva and Tadas Tem{\v{c}}inas and Daniela Gerz and
                  Matthew Henderson and Ivan Vuli{\'c}},
  title        = {Efficient Intent Detection with Dual Sentence Encoders},
  booktitle    = {Proceedings of the 2nd Workshop on Natural Language Processing for Conversational AI},
  pages        = {38--45},
  year         = {2020},
  publisher    = {Association for Computational Linguistics},
  note         = {Introduces the {BANKING77} dataset}
}

@article{clopper1934confidence,
  author       = {C. J. Clopper and E. S. Pearson},
  title        = {The Use of Confidence or Fiducial Limits Illustrated in the Case of the Binomial},
  journal      = {Biometrika},
  volume       = {26},
  number       = {4},
  pages        = {404--413},
  year         = {1934}
}

@inproceedings{fitzgerald2023massive,
  author       = {Jack FitzGerald and Christopher Hench and Charith Peris and Scott Mackie and
                  Kay Rottmann and Ana Sanchez and Aaron Nash and Liam Urbach and
                  Vishesh Kakarala and Richa Singh and Swetha Ranganath and Laurie Crist and
                  Misha Britan and Wouter Leeuwis and Gokhan Tur and Prem Natarajan},
  title        = {{MASSIVE}: A 1{M}-Example Multilingual Natural Language Understanding Dataset
                  with 51 Typologically-Diverse Languages},
  booktitle    = {Proceedings of the 61st Annual Meeting of the Association for Computational
                  Linguistics (Volume 1: Long Papers)},
  pages        = {4277--4302},
  year         = {2023},
  publisher    = {Association for Computational Linguistics}
}

@misc{granite2024granite,
  author       = {{Granite Team, IBM}},
  title        = {Granite 3.0 Language Models},
  year         = {2024},
  month        = oct,
  howpublished = {Technical report, \url{https://github.com/ibm-granite/granite-3.0-language-models/}}
}

@misc{grattafiori2024llama3,
  author       = {Aaron Grattafiori and Abhimanyu Dubey and Abhinav Jauhri and others},
  title        = {The Llama 3 Herd of Models},
  year         = {2024},
  eprint       = {2407.21783},
  archivePrefix = {arXiv},
  howpublished = {arXiv preprint arXiv:2407.21783}
}

@inproceedings{guo2017calibration,
  author       = {Chuan Guo and Geoff Pleiss and Yu Sun and Kilian Q. Weinberger},
  title        = {On Calibration of Modern Neural Networks},
  booktitle    = {Proceedings of the 34th International Conference on Machine Learning},
  series       = {Proceedings of Machine Learning Research},
  volume       = {70},
  pages        = {1321--1330},
  year         = {2017},
  publisher    = {PMLR}
}

@inproceedings{holtzman2021surface,
  author       = {Ari Holtzman and Peter West and Vered Shwartz and Yejin Choi and Luke Zettlemoyer},
  title        = {Surface Form Competition: Why the Highest Probability Answer Isn't Always Right},
  booktitle    = {Proceedings of the 2021 Conference on Empirical Methods in Natural Language Processing},
  pages        = {7038--7051},
  year         = {2021},
  publisher    = {Association for Computational Linguistics}
}

@misc{jiang2023mistral,
  author       = {Albert Q. Jiang and Alexandre Sablayrolles and Arthur Mensch and Chris Bamford and
                  Devendra Singh Chaplot and Diego de las Casas and Florian Bressand and
                  Gianna Lengyel and Guillaume Lample and Lucile Saulnier and
                  L{\'e}lio Renard Lavaud and Marie-Anne Lachaux and Pierre Stock and
                  Teven Le Scao and Thibaut Lavril and Thomas Wang and Timoth{\'e}e Lacroix and
                  William El Sayed},
  title        = {Mistral 7{B}},
  year         = {2023},
  eprint       = {2310.06825},
  archivePrefix = {arXiv},
  howpublished = {arXiv preprint arXiv:2310.06825}
}

@inproceedings{kwon2023pagedattention,
  author       = {Woosuk Kwon and Zhuohan Li and Siyuan Zhuang and Ying Sheng and Lianmin Zheng and
                  Cody Hao Yu and Joseph E. Gonzalez and Hao Zhang and Ion Stoica},
  title        = {Efficient Memory Management for Large Language Model Serving with {PagedAttention}},
  booktitle    = {Proceedings of the 29th Symposium on Operating Systems Principles (SOSP)},
  pages        = {611--626},
  year         = {2023},
  publisher    = {ACM}
}

@inproceedings{lang1995newsweeder,
  author       = {Ken Lang},
  title        = {{NewsWeeder}: Learning to Filter Netnews},
  booktitle    = {Machine Learning Proceedings 1995: Proceedings of the Twelfth International
                  Conference on Machine Learning},
  pages        = {331--339},
  year         = {1995},
  publisher    = {Morgan Kaufmann},
  note         = {Original source of the 20 Newsgroups dataset}
}

@misc{nostalgebraist2020logitlens,
  author       = {{nostalgebraist}},
  title        = {interpreting {GPT}: the logit lens},
  year         = {2020},
  month        = aug,
  howpublished = {LessWrong,
                  \url{https://www.lesswrong.com/posts/AcKRB8wDpdaN6v6ru/interpreting-gpt-the-logit-lens}},
  note         = {Published 31 August 2020}
}

@misc{olmo2024olmo2,
  author       = {{Team OLMo} and Pete Walsh and Luca Soldaini and Dirk Groeneveld and others},
  title        = {2 {OLMo} 2 Furious},
  year         = {2024},
  eprint       = {2501.00656},
  archivePrefix = {arXiv},
  howpublished = {arXiv preprint arXiv:2501.00656}
}

@misc{openai2025gptoss,
  author       = {{OpenAI}},
  title        = {gpt-oss-120b \& gpt-oss-20b Model Card},
  year         = {2025},
  eprint       = {2508.10925},
  archivePrefix = {arXiv},
  howpublished = {arXiv preprint arXiv:2508.10925}
}

@misc{qwen2024qwen25,
  author       = {{Qwen} and An Yang and Baosong Yang and Beichen Zhang and others},
  title        = {Qwen2.5 Technical Report},
  year         = {2024},
  eprint       = {2412.15115},
  archivePrefix = {arXiv},
  howpublished = {arXiv preprint arXiv:2412.15115}
}

@inproceedings{wolf2020transformers,
  author       = {Thomas Wolf and Lysandre Debut and Victor Sanh and Julien Chaumond and
                  Clement Delangue and Anthony Moi and Pierric Cistac and Tim Rault and
                  Remi Louf and Morgan Funtowicz and Joe Davison and Sam Shleifer and
                  Patrick von Platen and Clara Ma and Yacine Jernite and Julien Plu and
                  Canwen Xu and Teven Le Scao and Sylvain Gugger and Mariama Drame and
                  Quentin Lhoest and Alexander Rush},
  title        = {Transformers: State-of-the-Art Natural Language Processing},
  booktitle    = {Proceedings of the 2020 Conference on Empirical Methods in Natural Language
                  Processing: System Demonstrations},
  pages        = {38--45},
  year         = {2020},
  publisher    = {Association for Computational Linguistics}
}

@misc{yang2025qwen3,
  author       = {An Yang and Anfeng Li and Baosong Yang and Beichen Zhang and others},
  title        = {Qwen3 Technical Report},
  year         = {2025},
  eprint       = {2505.09388},
  archivePrefix = {arXiv},
  howpublished = {arXiv preprint arXiv:2505.09388}
}

@inproceedings{zhao2021calibrate,
  author       = {Zihao Zhao and Eric Wallace and Shi Feng and Dan Klein and Sameer Singh},
  title        = {Calibrate Before Use: Improving Few-shot Performance of Language Models},
  booktitle    = {Proceedings of the 38th International Conference on Machine Learning},
  series       = {Proceedings of Machine Learning Research},
  volume       = {139},
  pages        = {12697--12706},
  year         = {2021},
  publisher    = {PMLR}
}

@inproceedings{zheng2024robust,
  author       = {Chujie Zheng and Hao Zhou and Fandong Meng and Jie Zhou and Minlie Huang},
  title        = {Large Language Models Are Not Robust Multiple Choice Selectors},
  booktitle    = {International Conference on Learning Representations (ICLR)},
  year         = {2024}
}

@inproceedings{zhou2024batch,
  author       = {Han Zhou and Xingchen Wan and Lev Proleev and Diana Mincu and Jilin Chen and
                  Katherine Heller and Subhrajit Roy},
  title        = {Batch Calibration: Rethinking Calibration for In-Context Learning and Prompt
                  Engineering},
  booktitle    = {International Conference on Learning Representations (ICLR)},
  year         = {2024}
}

@misc{jevbench2026,
  author       = {Standhartinger, Florian},
  title        = {{JevBench}: a benchmark for typed decision models},
  year         = {2026},
  howpublished = {Hugging Face dataset, \url{https://huggingface.co/datasets/fstandhartinger/jevbench}},
  note         = {Public subset used here: 231 items in \texttt{datasets/public/}}
}

@misc{almeida2026jev,
  author       = {Diogo Almeida},
  title        = {Introducing {System One} Models \& {Jev}},
  year         = {2026},
  month        = sep,
  howpublished = {TypeSafe AI blog,
                  \url{https://typesafe.ai/blog/introducing-system-one-models-and-jev}},
  note         = {Published 15 September 2026}
}

@inproceedings{schick2021pet,
  author        = {Schick, Timo  and
      Sch{\"u}tze, Hinrich},
  title         = {{Exploiting Cloze-Questions for Few-Shot Text Classification and Natural Language Inference}},
  booktitle     = {Proceedings of the 16th Conference of the European Chapter of the Association for Computational Linguistics: Main Volume},
  year          = {2021},
  pages         = {255--269},
  publisher     = {Association for Computational Linguistics},
  doi           = {10.18653/v1/2021.eacl-main.20},
  url           = {https://aclanthology.org/2021.eacl-main.20/}
}

@inproceedings{gao2021lmbff,
  author        = {Gao, Tianyu  and
      Fisch, Adam  and
      Chen, Danqi},
  title         = {{Making Pre-trained Language Models Better Few-shot Learners}},
  booktitle     = {Proceedings of the 59th Annual Meeting of the Association for Computational Linguistics and the 11th International Joint Conference on Natural Language Processing (Volume 1: Long Papers)},
  year          = {2021},
  pages         = {3816--3830},
  publisher     = {Association for Computational Linguistics},
  doi           = {10.18653/v1/2021.acl-long.295},
  url           = {https://aclanthology.org/2021.acl-long.295/}
}

@misc{llm2jev,
  key          = {LLM2Jev},
  title        = {{LLM2Jev: Turn LLMs into Jev-Style Decision Models}},
  year         = {2026},
  howpublished = {GitHub repository, \url{https://github.com/Yinsongxu/LLM2Jev}},
  note         = {Accessed 29 September 2026}
}

@misc{llm2jev-joint,
  key          = {llm2jev-joint},
  title        = {{llm2jev}},
  year         = {2026},
  howpublished = {GitHub repository, \url{https://github.com/tic-top/llm2jev}},
  note         = {Accessed 29 September 2026}
}

@misc{semif,
  key           = {SemIf},
  title         = {{SemIf (formerly OpenJev)}},
  year          = {2026},
  howpublished  = {GitHub repository, \url{https://github.com/TheoLeeCJ/SemIf-OpenJev}},
  note          = {Accessed 29 September 2026}
}

@misc{nanojev,
  key          = {NanoJev},
  title        = {{NanoJev}: A Nano Replica of {Jev}},
  year         = {2026},
  howpublished = {GitHub repository, \url{https://github.com/TianyuCodings/NanoJev}},
  note         = {Accessed 29 September 2026}
}

@misc{kev,
  key          = {Kev},
  title        = {{Kev}},
  year         = {2026},
  howpublished = {GitHub repository, \url{https://github.com/jaredpalmer/kev}},
  note         = {Accessed 29 September 2026}
}

@misc{nimble,
  author       = {{Bespoke Labs} and Sathiamoorthy, Maheswaran},
  title        = {{Nimble}},
  year         = {2026},
  howpublished = {GitHub repository, \url{https://github.com/bespokelabsai/nimble}},
  note         = {Accessed 29 September 2026}
}

@misc{laya,
  key          = {Laya},
  title        = {{Laya}: Multilingual, Non-Autoregressive System 1 Decision Engine},
  year         = {2026},
  howpublished = {GitHub repository, \url{https://github.com/NandhaKishorM/laya}},
  note         = {Accessed 29 September 2026}
}

@misc{decider,
  author = {Marosi, Mark},
  title  = {decider: one-pass typed decisions with calibrated probabilities},
  year   = {2026},
  howpublished = {GitHub repository, \url{https://github.com/Mapika/decider}},
  url    = {https://github.com/Mapika/decider}
}

@misc{jevlike,
  key          = {Jevlike},
  title        = {{Jevlike}},
  year         = {2026},
  howpublished = {GitHub repository, \url{https://github.com/vinnylarouge/jevlike}},
  note         = {Accessed 29 September 2026}
}

@misc{jevos,
  author       = {{feder-cr}},
  title        = {{jevos}},
  year         = {2026},
  howpublished = {GitHub repository, \url{https://github.com/feder-cr/jev}},
  note         = {Accessed 29 September 2026}
}

@misc{yu2026visualjev,
  author        = {Guanxu Yu and Yuhang Yao},
  title         = {{Visual Jev: Accurate and Efficient Decisions from Shared Visual Context}},
  year          = {2026},
  eprint        = {2609.25845},
  archivePrefix = {arXiv},
  howpublished  = {arXiv preprint arXiv:2609.25845},
  url           = {https://arxiv.org/abs/2609.25845},
  note          = {Code: \url{https://github.com/guanxuyu-sv/Visual-Jev}}
}

@misc{ye2026numericjev,
  author        = {Weiwei Ye and Hangchen Liu and Renhe Jiang},
  title         = {{NumericJev: Jev-like LLM Numerical Decoding with Multiway Decision Trees}},
  year          = {2026},
  eprint        = {2609.28587},
  archivePrefix = {arXiv},
  howpublished  = {arXiv preprint arXiv:2609.28587},
  url           = {https://arxiv.org/abs/2609.28587}
}

@misc{sun2026typesafe,
  author        = {Yu Sun and Junhao Xu and Jiajia Shi and Zijin Yang},
  title         = {{Type-Safe Is Not Error-Free: A Constrained Decision Head Follows the Option Name, Not the Rubric Bound to It}},
  year          = {2026},
  eprint        = {2609.26758},
  archivePrefix = {arXiv},
  howpublished  = {arXiv preprint arXiv:2609.26758},
  url           = {https://arxiv.org/abs/2609.26758}
}

@misc{li2026coherence,
  author        = {Keyi Li and Yihao He and Quanyi Li},
  title         = {{Beyond Calibration: Do a Typed-Decision Model's Probabilities Obey the Probability Axioms?}},
  year          = {2026},
  eprint        = {2609.33209},
  archivePrefix = {arXiv},
  howpublished  = {arXiv preprint arXiv:2609.33209},
  url           = {https://arxiv.org/abs/2609.33209},
  note          = {Code: \url{https://github.com/bro789/typed-decision-coherence}}
}

@article{chen2024frugalgpt,
  author        = {Lingjiao Chen and Matei Zaharia and James Zou},
  title         = {{FrugalGPT: How to Use Large Language Models While Reducing Cost and Improving Performance}},
  journal       = {Transactions on Machine Learning Research},
  year          = {2024},
  url           = {https://arxiv.org/abs/2305.05176},
  note          = {Code: \url{https://github.com/stanford-futuredata/FrugalGPT}}
}

@misc{wu2026reflex,
  author        = {Tiantong Wu and Wei Yang Bryan Lim},
  title         = {{REFLEX with Jev for Efficient Selective Control in LLM Agents}},
  year          = {2026},
  eprint        = {2609.26532},
  archivePrefix = {arXiv},
  howpublished  = {arXiv preprint arXiv:2609.26532},
  url           = {https://arxiv.org/abs/2609.26532}
}

@misc{li2026jevjudge,
  author        = {Yubo Li and Yidi Miao and Ramayya Krishnan and Rema Padman},
  title         = {{JEV-as-a-Judge: Accept When Confident, Escalate When Unsure}},
  year          = {2026},
  eprint        = {2609.26550},
  archivePrefix = {arXiv},
  howpublished  = {arXiv preprint arXiv:2609.26550},
  url           = {https://arxiv.org/abs/2609.26550}
}

@misc{rao2026rubricjudges,
  author        = {Delip Rao and Chris Callison-Burch},
  title         = {{JEV vs. LLMs as Rubric Judges: Cheaper, Faster, and Wrong in the Same Places}},
  year          = {2026},
  eprint        = {2609.29769},
  archivePrefix = {arXiv},
  howpublished  = {arXiv preprint arXiv:2609.29769},
  url           = {https://arxiv.org/abs/2609.29769}
}

@misc{jev-ultrafast,
  key          = {jev-ultrafast},
  title        = {{Jev Ultrafast}},
  year         = {2026},
  howpublished = {GitHub repository, \url{https://github.com/browser-use/jev-ultrafast}},
  note         = {Accessed 29 September 2026}
}

@misc{fast-jev-compaction,
  key          = {fast-jev-compaction},
  title        = {{fast-jev-compaction}},
  year         = {2026},
  howpublished = {GitHub repository, \url{https://github.com/tamaratran/fast-jev-compaction}},
  note         = {Accessed 29 September 2026}
}

@misc{jevgrep,
  author       = {{dzhng}},
  title        = {{jevgrep}},
  year         = {2026},
  howpublished = {GitHub repository, \url{https://github.com/dzhng/jevgrep}},
  note         = {Accessed 29 September 2026}
}

@misc{zhang2026spectra,
  author       = {Jiamu Zhang and Liang Wu and Kelly Wan and Hanjie Chen and Liangjie Hong},
  title        = {{SPECTRA}: Pushing the {KV} Cache Beyond the 2-Bit Cliff via
                  Spectral Transform Coding},
  year         = {2026},
  eprint       = {2608.07915},
  archivePrefix = {arXiv},
  howpublished = {arXiv preprint arXiv:2608.07915},
  url          = {https://arxiv.org/abs/2608.07915},
  note         = {Code: \url{https://github.com/nokia-applied-research/SPECTRA}}
}

@misc{min2026trace,
  author       = {Guanghui Min and Liang Wu and Mayank Darbari and Chen Chen and Liangjie Hong},
  title        = {Toward Reliable Context Compression for Long-Horizon Agents:
                  An Empirical Study of Execution Instability},
  year         = {2026},
  eprint       = {2608.06503},
  archivePrefix = {arXiv},
  howpublished = {arXiv preprint arXiv:2608.06503},
  url          = {https://arxiv.org/abs/2608.06503},
  note         = {Code: \url{https://github.com/nokia-applied-research/Trace}}
}

@misc{zhang2026wisp,
  author       = {Jiamu Zhang and Liang Wu and Mayank Darbari and Liangjie Hong},
  title        = {{WiSP}: A Working-Set View of Mixture-of-Experts Serving on
                  Extremely Low-Resource Hardware},
  year         = {2026},
  eprint       = {2606.21868},
  archivePrefix = {arXiv},
  howpublished = {arXiv preprint arXiv:2606.21868},
  url          = {https://arxiv.org/abs/2606.21868},
  note         = {Code: \url{https://github.com/nokia-applied-research/WiSP}}
}

@misc{wu2026autonomousnetworks,
  author       = {Liang Wu and Kelly Wan and Mayank Darbari and Liangjie Hong},
  title        = {Towards Resilient and Autonomous Networks: A {BlueSky} Vision
                  on {AI}-Native {6G}},
  year         = {2026},
  eprint       = {2605.21395},
  archivePrefix = {arXiv},
  howpublished = {arXiv preprint arXiv:2605.21395},
  url          = {https://arxiv.org/abs/2605.21395},
  note         = {Accepted at KDD 2026}
}

@misc{li2026sixgorchestration,
  author        = {Delong Li and Xu Wang and Haochen Gong and Rui Lang and Guangsheng Yu},
  title         = {{Fast Intent-Driven Service Orchestration with Jev for 6G Edge Networks}},
  year          = {2026},
  eprint        = {2609.23136},
  archivePrefix = {arXiv},
  howpublished  = {arXiv preprint arXiv:2609.23136},
  url           = {https://arxiv.org/abs/2609.23136}
}

@misc{yang2026codewithimage,
  author       = {Tianze Yang and Liang Wu and Ruitong Sun and Yucheng Shi and
                  Yanqiao Wang and Mayank Darbari and Ninghao Liu and Jin Sun and Liangjie Hong},
  title        = {Self-Evolving Code-with-Image Reasoning},
  year         = {2026},
  eprint       = {2608.11292},
  archivePrefix = {arXiv},
  howpublished = {arXiv preprint arXiv:2608.11292},
  url          = {https://arxiv.org/abs/2608.11292}
}

@misc{goulart2026fieldaware,
  author       = {Paimon Goulart and Liang Wu and Kelly Wan and
                  Evangelos E. Papalexakis and Liangjie Hong},
  title        = {Field-Aware Agent Skill Retrieval},
  year         = {2026},
  eprint       = {2608.02880},
  archivePrefix = {arXiv},
  howpublished = {arXiv preprint arXiv:2608.02880},
  url          = {https://arxiv.org/abs/2608.02880}
}

@misc{wang2026pagellm,
  author       = {Xinyuan Wang and Liang Wu and Dongjie Wang and Yanjie Fu},
  title        = {{PageLLM}: A Multi-Grained Reward Framework for Whole-Page
                  Optimization with Large Language Models},
  year         = {2026},
  eprint       = {2506.09084},
  archivePrefix = {arXiv},
  howpublished = {arXiv preprint arXiv:2506.09084},
  url          = {https://arxiv.org/abs/2506.09084}
}

@misc{zheng2026sapo,
  author       = {Zaiyi Zheng and Liang Wu and Guanghui Min and Yaochen Zhu and
                  Liangjie Hong and Chen Chen and Jundong Li},
  title        = {{SAPO}: Step-Aligned Policy Optimization for Reasoning-Based
                  Generative Recommendation},
  year         = {2026},
  eprint       = {2605.17648},
  archivePrefix = {arXiv},
  howpublished = {arXiv preprint arXiv:2605.17648},
  url          = {https://arxiv.org/abs/2605.17648}
}

@misc{chen2026timeseries,
  author       = {Yilong Chen and Xiao Qin and Chenghao Liu and Liang Wu and
                  Noelle I. Samia and Kaize Ding},
  title        = {{LLM} Agents for Time-Series: A Survey},
  year         = {2026},
  eprint       = {2608.26226},
  archivePrefix = {arXiv},
  howpublished = {arXiv preprint arXiv:2608.26226},
  url          = {https://arxiv.org/abs/2608.26226},
  note         = {Accepted to Findings of EMNLP 2026}
}

@misc{bevdecision2026,
  author       = {{avbiswas}},
  title        = {{BEV Decision Mix}},
  year         = {2026},
  howpublished = {Hugging Face dataset, \url{https://huggingface.co/datasets/avbiswas/bev-decision}},
  note         = {Default configuration; accessed September 30, 2026}
}

\appendix
\section{Per-model results on the 20-option tasks}
\label{app:breadth}

Table~\ref{tab:breadth} reports means. The three tables below give each model separately.
\emph{mass} is the answer mass of Eq.~\eqref{eq:restrict}, in bold where it falls below the 0.99
gate. \emph{rot.}\ is the rotation average of Eq.~\eqref{eq:rot} with the label prior switched off,
\emph{prior} adds Eq.~\eqref{eq:prior}, and \emph{temp.}\ adds the fitted temperature.
\emph{cov@5\%} is the share of items the system can answer, taken in confidence order, before the
error rate on the answered set passes 5\%.

\begin{table}[h]
  \centering
  \scriptsize\setlength{\tabcolsep}{4pt}
\begin{tabular}{l c ccc cc cc cc}
\toprule
Model & mass & raw & +rot. & +prior & ECE raw & ECE +temp. & flip raw & flip +rot. & cov@5\% raw & cov@5\% +temp. \\
\midrule
Granite-3.3-8B & 1.000 & 0.657 & 0.810 & 0.807 & 0.293 & 0.052 & 0.353 & 0.143 & 0.033 & 0.523 \\
Qwen3-8B & 1.000 & 0.747 & 0.800 & 0.803 & 0.240 & 0.095 & 0.230 & 0.077 & 0.077 & 0.520 \\
Phi-4-mini & 0.999 & 0.707 & 0.767 & 0.783 & 0.185 & 0.065 & 0.277 & 0.130 & 0.017 & 0.480 \\
Qwen3-32B & 1.000 & 0.720 & 0.777 & 0.780 & 0.227 & 0.064 & 0.213 & 0.080 & 0.017 & 0.443 \\
Qwen3-30B-A3B & 1.000 & 0.730 & 0.757 & 0.770 & 0.249 & 0.086 & 0.143 & 0.103 & 0.277 & 0.470 \\
Qwen3-4B & 1.000 & 0.733 & 0.757 & 0.760 & 0.254 & 0.098 & 0.197 & 0.110 & 0.103 & 0.423 \\
Qwen2.5-7B & 0.998 & 0.723 & 0.750 & 0.757 & 0.237 & 0.070 & 0.197 & 0.073 & 0.180 & 0.223 \\
Mistral-7B & 1.000 & 0.693 & 0.727 & 0.740 & 0.287 & 0.113 & 0.290 & 0.137 & 0.100 & 0.347 \\
SmolLM2-1.7B & 0.992 & 0.407 & 0.667 & 0.717 & 0.219 & 0.102 & 0.867 & 0.263 & 0.093 & 0.360 \\
OLMo-2-7B & 0.999 & 0.600 & 0.633 & 0.680 & 0.185 & 0.105 & 0.403 & 0.203 & 0.003 & 0.037 \\
Qwen3-1.7B & 1.000 & 0.597 & 0.667 & 0.670 & 0.383 & 0.121 & 0.433 & 0.197 & 0.027 & 0.260 \\
\bottomrule
\end{tabular}

  \caption{\texttt{banking20}, $K=20$, 300 items per model. All eleven rows pass the gate.}
\end{table}

\begin{table}[h]
  \centering
  \scriptsize\setlength{\tabcolsep}{4pt}
\begin{tabular}{l c ccc cc cc cc}
\toprule
Model & mass & raw & +rot. & +prior & ECE raw & ECE +temp. & flip raw & flip +rot. & cov@5\% raw & cov@5\% +temp. \\
\midrule
Qwen3-30B-A3B & 1.000 & 0.737 & 0.743 & 0.740 & 0.242 & 0.096 & 0.140 & 0.087 & 0.303 & 0.003 \\
Qwen3-32B & 1.000 & 0.717 & 0.733 & 0.737 & 0.236 & 0.063 & 0.170 & 0.090 & 0.557 & 0.410 \\
Qwen2.5-7B & 0.998 & 0.663 & 0.703 & 0.710 & 0.273 & 0.096 & 0.237 & 0.133 & 0.047 & 0.407 \\
Granite-3.3-8B & 1.000 & 0.660 & 0.690 & 0.693 & 0.328 & 0.111 & 0.303 & 0.187 & 0.000 & 0.183 \\
OLMo-2-7B & 0.999 & 0.573 & 0.660 & 0.683 & 0.203 & 0.047 & 0.457 & 0.240 & 0.283 & 0.290 \\
Mistral-7B & 0.995 & 0.647 & 0.663 & 0.673 & 0.315 & 0.096 & 0.317 & 0.167 & 0.007 & 0.073 \\
Qwen3-4B & 1.000 & 0.633 & 0.667 & 0.667 & 0.341 & 0.138 & 0.273 & 0.200 & 0.000 & 0.157 \\
Qwen3-8B & 0.999 & 0.637 & 0.660 & 0.660 & 0.334 & 0.138 & 0.233 & 0.173 & 0.013 & 0.237 \\
SmolLM2-1.7B & 0.994 & 0.257 & 0.610 & 0.637 & 0.097 & 0.089 & 0.920 & 0.243 & 0.003 & 0.137 \\
Phi-4-mini & 0.999 & 0.593 & 0.613 & 0.617 & 0.272 & 0.110 & 0.310 & 0.233 & 0.080 & 0.217 \\
Qwen3-1.7B & 1.000 & 0.580 & 0.603 & 0.610 & 0.394 & 0.106 & 0.313 & 0.257 & 0.010 & 0.000 \\
\bottomrule
\end{tabular}

  \caption{\texttt{newsgroups}, $K=20$, 300 items per model. All eleven rows pass the gate.}
\end{table}

\begin{table}[h]
  \centering
  \scriptsize\setlength{\tabcolsep}{4pt}
\begin{tabular}{l c ccc cc cc cc}
\toprule
Model & mass & raw & +rot. & +prior & ECE raw & ECE +temp. & flip raw & flip +rot. & cov@5\% raw & cov@5\% +temp. \\
\midrule
Qwen3-32B & 0.999 & 0.857 & 0.833 & 0.827 & 0.062 & 0.067 & 0.073 & 0.000 & 0.550 & 0.483 \\
Phi-4-mini & 0.994 & 0.707 & 0.790 & 0.807 & 0.133 & 0.106 & 0.427 & 0.000 & 0.017 & 0.010 \\
Qwen3-4B & 1.000 & 0.683 & 0.783 & 0.797 & 0.298 & 0.081 & 0.203 & 0.000 & 0.117 & 0.100 \\
Qwen2.5-7B & 0.999 & 0.737 & 0.720 & 0.790 & 0.188 & 0.049 & 0.070 & 0.000 & 0.367 & 0.380 \\
Qwen3-30B-A3B & 0.995 & 0.730 & 0.730 & 0.757 & 0.248 & 0.090 & 0.103 & 0.000 & 0.410 & 0.410 \\
Mistral-7B & 0.995 & 0.697 & 0.707 & 0.733 & 0.167 & 0.069 & 0.333 & 0.000 & 0.017 & 0.143 \\
Qwen3-8B & 1.000 & 0.693 & 0.677 & 0.700 & 0.288 & 0.161 & 0.060 & 0.000 & 0.313 & 0.313 \\
Qwen3-1.7B & 0.992 & 0.657 & 0.627 & 0.683 & 0.231 & 0.135 & 0.170 & 0.000 & 0.000 & 0.000 \\
SmolLM2-1.7B & \textbf{0.408} & 0.593 & 0.593 & 0.670 & 0.215 & 0.095 & 0.000 & 0.000 & 0.030 & 0.000 \\
Granite-3.3-8B & \textbf{0.976} & 0.627 & 0.637 & 0.647 & 0.302 & 0.103 & 0.100 & 0.000 & 0.000 & 0.003 \\
OLMo-2-7B & \textbf{0.710} & 0.593 & 0.593 & 0.593 & 0.367 & 0.083 & 0.003 & 0.000 & 0.000 & 0.000 \\
\bottomrule
\end{tabular}

  \caption{\texttt{injection}, $K=2$, 300 items per model. Three rows fall below the answer-mass
  gate and are excluded from every statistic reported for this task. The rotation average removes
  order dependence on every row that passes.}
\end{table}

\Needspace{12\baselineskip}
\section{Benchmark detail}
\label{app:jevbench}

\begin{table}[h]
  \centering
  \scriptsize
\begin{tabular}{l ccc l ccc l}
\toprule
Model & raw & debiased & diff. & 95\% CI & ECE deb. & ECE +temp. & diff. & 95\% CI \\
\midrule
Qwen3-32B & 0.798 & 0.789 & -0.009 & [-0.042, +0.019] & 0.128 & 0.091 & -0.037 & [-0.082, +0.010] \\
Qwen3-8B & 0.704 & 0.714 & +0.009 & [-0.009, +0.028] & 0.263 & 0.114 & -0.149 & [-0.179, -0.076] \\
gpt-oss-20b & 0.704 & 0.714 & +0.009 & [-0.033, +0.052] & 0.154 & 0.102 & -0.052 & [-0.092, +0.004] \\
Granite-3.3-8B & 0.690 & 0.700 & +0.009 & [-0.019, +0.038] & 0.270 & 0.301 & +0.030 & [-0.038, +0.124] \\
Qwen2.5-7B & 0.676 & 0.685 & +0.009 & [-0.009, +0.033] & 0.243 & 0.085 & -0.159 & [-0.177, -0.068] \\
Mistral-7B & 0.615 & 0.624 & +0.009 & [-0.019, +0.038] & 0.333 & 0.104 & -0.229 & [-0.251, -0.121] \\
Qwen3-1.7B & 0.549 & 0.568 & +0.019 & [-0.023, +0.066] & 0.380 & 0.105 & -0.276 & [-0.297, -0.179] \\
Llama-3.2-3B & 0.563 & 0.559 & -0.005 & [-0.042, +0.033] & 0.163 & 0.107 & -0.057 & [-0.093, +0.019] \\
\bottomrule
\end{tabular}

  \caption{Per-model differences with 95\% bootstrap intervals over 2\,000 resamples, paired over
  items. The temperature is refitted inside every resample. Reusing a fit made on the full sample
  would place the whole sample inside every draw and narrow the interval.}
  \label{tab:jevbench-ci}
\end{table}

\begin{table}[h]
  \centering
  \small
\begin{tabular}{c cc cc c l}
\toprule
Options $K$ & Items & Decisions & Raw & Debiased & Difference & 95\% CI \\
\midrule
2 & 74 & 592 & 0.677 & 0.676 & -0.002 & [-0.017, +0.015] \\
3 & 15 & 120 & 0.333 & 0.317 & -0.017 & [-0.050, +0.017] \\
4 & 53 & 424 & 0.649 & 0.665 & +0.017 & [-0.012, +0.045] \\
5 & 55 & 440 & 0.805 & 0.811 & +0.007 & [-0.016, +0.030] \\
6 & 16 & 128 & 0.461 & 0.492 & +0.031 & [+0.008, +0.062] \\
\midrule
\multicolumn{5}{l}{Slope of the difference on $K$} & +0.0059 & [-0.0012, +0.0132] \\
\multicolumn{5}{l}{Pooled over all 1704 decisions} & +0.0065 & [-0.0041, +0.0176] \\
\bottomrule
\end{tabular}

  \caption{The difference against the number of options, all eight models pooled. Resampling is
  clustered by item, so a drawn item contributes all eight models' decisions. The final two rows replace those five interval tests with a single test of the trend.}
  \label{tab:by-k}
\end{table}

\begin{table}[h]
  \centering
  \small
\begin{tabular}{l ccc}
\toprule
Model & easy (48) & original (72) & hard (111) \\
\midrule
Qwen3-32B & 1.000 & 0.967 & 0.590 \\
Qwen3-8B & 1.000 & 0.817 & 0.524 \\
gpt-oss-20b & 1.000 & 0.850 & 0.505 \\
Granite-3.3-8B & 1.000 & 0.900 & 0.448 \\
Qwen2.5-7B & 1.000 & 0.867 & 0.438 \\
Mistral-7B & 1.000 & 0.717 & 0.400 \\
Llama-3.2-3B & 0.917 & 0.650 & 0.343 \\
Qwen3-1.7B & 1.000 & 0.600 & 0.352 \\
\bottomrule
\end{tabular}

  \caption{Accuracy by difficulty tier of the public subset. These are the three public files, not
  the partition the benchmark's overall score uses. The counts in the header are the published items
  per file; the accuracies are over the scored items only, which are 48 easy, 60 original and 105
  hard, because the 18 unscored items fall in the second and third files.}
\end{table}

\section{Rotations and the certificate}
\label{app:rotation}

\begin{figure}[h]
  \centering
  \includegraphics[width=\textwidth]{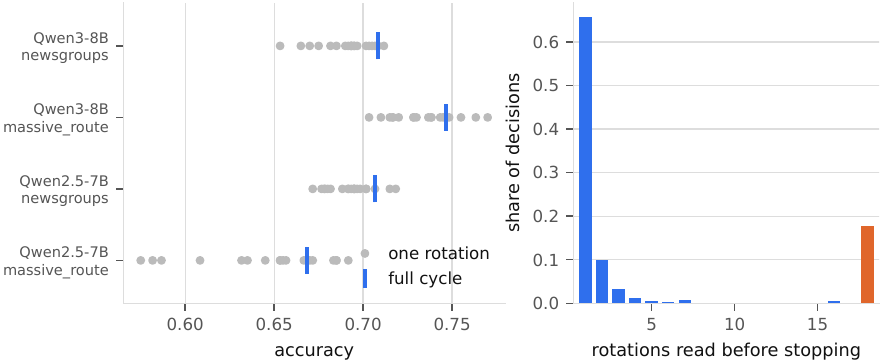}
  \caption{Left: every cyclic rotation scored on its own, against the full-rotation average, on four
  (model, task) cells. On all four the average falls at or below the best single
  rotation, by 0.3 to 2.3 points, and above the worst by 3.5 to 9.3 points. Right: where the budget stops on one 18-option cell, over 600
  decisions, for a rule with no minimum-rotation floor; the shipped default reads two rotations
  before it may stop, which moves the leftmost bar.}
  \label{fig:rotation}
\end{figure}

The left panel states what the rotations do. They do not raise accuracy above the best single
rotation. They remove the caller's ability to select a poor
one, and they reduce how far the result depends on the order the options were written in.

\begin{table}[h]
  \centering
  \small
\begin{tabular}{l l c c cc c cc}
\toprule
Model & Task & $K$ & Threshold & Verify rate & Verify bound & Clears 1\% & Rotations & vs.\ $K$ \\
\midrule
Qwen2.5-7B & massive\_route & 18 & 10.25 & 0.0017 & 0.0079 & yes & 10.61 & 1.70$\times$ \\
Qwen2.5-7B & newsgroups & 20 & 9.50 & 0.0017 & 0.0079 & yes & 9.01 & 2.22$\times$ \\
Qwen3-8B & massive\_route & 18 & --- & --- & --- & --- & --- & --- \\
Qwen3-8B & newsgroups & 20 & 9.00 & 0.0033 & 0.0105 & no & 5.31 & 3.77$\times$ \\
\bottomrule
\end{tabular}

  \caption{The certificate with the threshold selected on one third of an unlabelled batch and the
  bound computed on the other two thirds, for that one threshold. The verify bound carries no
  selection multiplicity. Two of the four cells clear a 1\% target this way; on
  \texttt{massive\_route} with Qwen3-8B no threshold on the grid cleared it on the selection split.}
  \label{tab:cert-split}
\end{table}

\begin{table}[h]
  \centering
  \small
\begin{tabular}{l cc}
\toprule
Stopping statistic & Cells certified & Mean rotations saved \\
\midrule
log-odds margin alone & 4 of 4 & 3.46$\times$ \\
log-odds margin + unanimity & 4 of 4 & 2.91$\times$ \\
probability gap alone & 2 of 4 & 3.19$\times$ \\
probability gap + unanimity & 2 of 4 & 2.64$\times$ \\
\bottomrule
\end{tabular}

  \caption{Which stopping statistic satisfies Eq.~\eqref{eq:cert} at $\varepsilon = 0.01$, over
  four (model, task) cells. A rule with fewer certified cells did not run worse: no threshold on the
  grid cleared the target on those cells at any budget.}
  \label{tab:certificate}
\end{table}

\begin{table}[h]
  \centering
  \footnotesize
\begin{tabular}{l l l c c ccc}
\toprule
Stopping statistic & Model & Task & $K$ & Threshold & Rotations & vs.\ $K$ & Disagreement \\
\midrule
log-odds margin alone & Qwen2.5-7B & massive\_route & 18 & 4.5 & 6.12 & 2.94$\times$ & 0.0083 \\
 & Qwen2.5-7B & newsgroups & 20 & 6.0 & 6.00 & 3.34$\times$ & 0.0000 \\
 & Qwen3-8B & massive\_route & 18 & 9.5 & 5.37 & 3.36$\times$ & 0.0000 \\
 & Qwen3-8B & newsgroups & 20 & 8.25 & 4.75 & 4.21$\times$ & 0.0033 \\
log-odds margin + unanimity & Qwen2.5-7B & massive\_route & 18 & 4.25 & 7.96 & 2.26$\times$ & 0.0083 \\
 & Qwen2.5-7B & newsgroups & 20 & 6.0 & 6.63 & 3.02$\times$ & 0.0000 \\
 & Qwen3-8B & massive\_route & 18 & 9.5 & 6.38 & 2.82$\times$ & 0.0000 \\
 & Qwen3-8B & newsgroups & 20 & 8.25 & 5.63 & 3.55$\times$ & 0.0017 \\
probability gap alone & Qwen2.5-7B & massive\_route & 18 & 0.97 & 5.87 & 3.07$\times$ & 0.0083 \\
 & Qwen2.5-7B & newsgroups & 20 & 0.995 & 6.05 & 3.30$\times$ & 0.0000 \\
probability gap + unanimity & Qwen2.5-7B & massive\_route & 18 & 0.965 & 7.89 & 2.28$\times$ & 0.0083 \\
 & Qwen2.5-7B & newsgroups & 20 & 0.995 & 6.69 & 2.99$\times$ & 0.0000 \\
\bottomrule
\end{tabular}

  \label{tab:certificate-cells}
  \caption{Every cell each stopping rule certifies at $\varepsilon = 0.01$, with its threshold and
  the rotations it then reads. Rules with two rows did not certify the other two cells at any
  threshold on the grid. These thresholds come from a different calibration sample than the serving
  run of Section~\ref{sec:results}, which is why the same cell appears there at 6.0.}
\end{table}

\section{Reading a truncated forward}
\label{app:exit}

Truncating the forward at block $b$ leaves the residual stream outside the basis that the output
head reads. We fit an affine map from block $b$ into that basis, targeting the model's own
full-depth state on unlabelled inputs, which is a tuned lens~\cite{belrose2023tuned}. We select the
depth on a calibration corpus disjoint from every evaluation set. Reading the truncated state
through the output head without the map is the logit lens~\cite{nostalgebraist2020logitlens}. The
quantity throughout is agreement with the model's own full-depth decision, not accuracy.

Across the eight models the transition is sharp. Seven never exceed 0.47 agreement at any depth
below the crossing, and then reach 0.90 in a single measured step. The depth at which that happens
ranges from 66\% to 100\% of blocks and does not follow parameter count. At a 0.98 target six of the
eight models save no blocks. The two that do are Mistral-7B at 66\% of blocks and Qwen3-8B at 89\%.

Part of what the table charges to truncation is the map's own error. At full depth no truncation
occurs, so any agreement below 1.000 in the last column is the map alone. It is 0.000 on Mistral-7B and 0.007 on Qwen3-8B, the two models that save blocks, and 0.040 on Llama-3.2-3B, which is why that model reaches no depth at the 0.98 target.

\begin{table}[h]
  \centering
  \footnotesize
\begin{tabular}{l c cc ll c}
\toprule
Model & Blocks & Depth at 0.98 & Depth at 0.95 & Before crossing & At crossing & Agreement at full depth \\
\midrule
Mistral-7B & 32 & 66\% & 66\% & 0.320 at 59\% & 0.985 at 66\% & 1.000 \\
Qwen3-8B & 36 & 89\% & 69\% & 0.087 at 64\% & 0.965 at 69\% & 0.993 \\
Qwen3-1.7B & 28 & 100\% & 79\% & 0.188 at 75\% & 0.958 at 79\% & 0.990 \\
Qwen2.5-7B & 28 & 100\% & 89\% & 0.420 at 79\% & 0.950 at 89\% & 0.983 \\
Qwen3-32B & 64 & 100\% & 91\% & 0.228 at 80\% & 0.973 at 91\% & 0.990 \\
gpt-oss-20b & 24 & 100\% & 92\% & 0.207 at 79\% & 0.950 at 92\% & 0.983 \\
Granite-3.3-8B & 40 & 100\% & 100\% & 0.193 at 80\% & 0.915 at 90\% & 0.995 \\
Llama-3.2-3B & 28 & not reached & 100\% & 0.770 at 89\% & 0.960 at 100\% & 0.960 \\
\bottomrule
\end{tabular}

  \caption{Agreement with the model's own full-depth decision, for a readout taken at a fraction of
  the blocks and mapped into the final basis. \emph{Before crossing} and \emph{at crossing} are the
  last depth measured below 0.90 agreement and the first at or above it.}
  \label{tab:layerskip}
\end{table}

\begin{figure}[h]
  \centering
  \includegraphics[width=\textwidth]{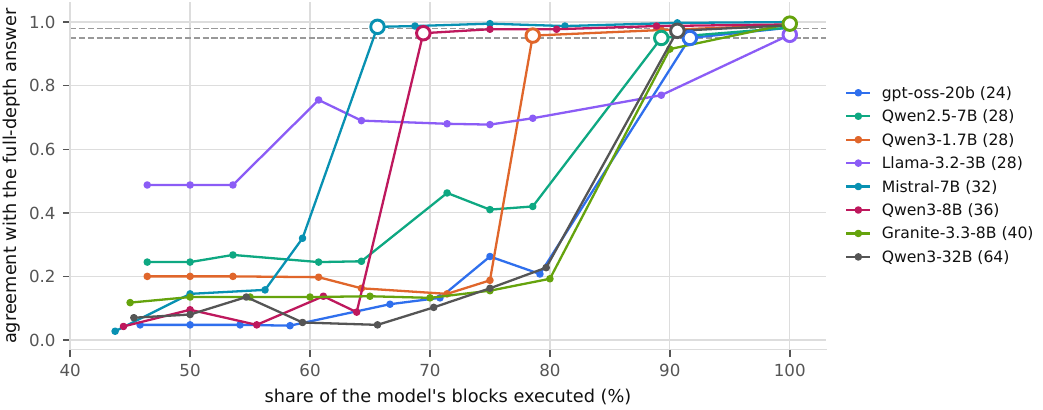}
  \caption{Agreement against depth, eight models. Markers are the depths we ran, and the curve is
  drawn through them rather than smoothed, because the intermediate depths were not measured. Hollow
  markers show the first depth reaching 0.95. Dashed lines mark 0.95 and 0.98.}
  \label{fig:layerskip}
\end{figure}

\clearpage
\section{Full BEV Decision Mix results}
\label{app:bev-full}

We evaluate the original \texttt{default} configuration of BEV Decision Mix~\cite{bevdecision2026}.
Its 24\,386 test rows expand to 46\,320 state--question decisions: 18\,808 choices among 2--10
options, 17\,572 yes/no decisions (\texttt{noul}), and 9\,940 ordinal scores with 3--7 levels.
Table~\ref{tab:bev-full} summarizes all readouts evaluated on the full test split.
The sampled Qwen2.5-7B depth experiment is excluded from this comparison.

\begin{table}[!ht]
  \centering
  \caption{All full-split BEV Decision Mix results. Accuracy and its intervals are in percent;
  ECE is on a 0--1 scale. Higher accuracy and lower ECE are better.
  Bold marks the best measured system in each metric, including ties.}
  \label{tab:bev-full}
  \begingroup
\fontsize{9}{11}\selectfont
\setlength{\tabcolsep}{3.4pt}
\renewcommand{\arraystretch}{1.06}
\begin{tabular*}{\textwidth}{@{\extracolsep{\fill}}l r c r r r r@{}}
\toprule
System / readout & Overall $\uparrow$ & 95\% CI & Choice & Yes/no & Ordinal & ECE $\downarrow$ \\
 & & & $n=18{,}808$ & $n=17{,}572$ & $n=9{,}940$ & (shard mean) \\
\midrule
\addlinespace[3pt]
\multicolumn{7}{@{}l}{\textit{Reference rules (not learned systems)}} \\
Uniform random (expected) & 34.29 & --- & 25.77 & 50.00 & 22.63 & --- \\
Majority by question (test oracle) & 54.28 & --- & 41.42 & 73.57 & 44.51 & --- \\
\midrule
\addlinespace[3pt]
\multicolumn{7}{@{}l}{\textit{Published baseline interfaces}} \\
Bespoke-Nimble-9B & \textbf{70.57} & [70.16, 70.99] & \textbf{73.60} & 81.06 & 46.33 & 0.0889 \\
Laya typed-decisions & 56.20 & [55.75, 56.65] & 55.94 & 72.97 & 27.05 & 0.0444 \\
Qwen3-Reranker-0.6B & 40.50 & [40.05, 40.95] & 52.47 & 37.46 & 23.22 & 0.2249 \\
NanoJev & 32.99 & [32.56, 33.42] & 35.43 & 34.57 & 25.55 & 0.2151 \\
\midrule
\addlinespace[3pt]
\multicolumn{7}{@{}l}{\textit{AnyJev on Qwen3.5-9B}} \\
Raw (one forward) & 67.85 & [67.43, 68.28] & 67.46 & 81.20 & 45.00 & 0.0699 \\
L0: rotations, prior off & 68.01 & [67.59, 68.43] & 67.69 & \textbf{81.37} & 45.00 & 0.0490 \\
L0: batch prior, $\alpha=0.75$ & 65.01 & [64.58, 65.45] & 66.37 & 74.60 & 45.50 & \textbf{0.0418} \\
Rotations + batch prior, $\alpha=1.0$ & 62.21 & [61.77, 62.65] & 63.70 & 70.69 & 44.42 & 0.0565 \\
Rotations + content-free, $\alpha=0.5$ & 64.45 & [64.01, 64.88] & 69.39 & 71.86 & 41.99 & 0.0839 \\
Rotations + content-free, $\alpha=0.75$ & 62.10 & [61.66, 62.54] & 69.52 & 66.38 & 40.48 & 0.1122 \\
Rotations + content-free, $\alpha=1.0$ & 58.95 & [58.50, 59.40] & 68.55 & 60.68 & 37.75 & 0.1483 \\
L2-mono (affine map, block 29/32) & 67.24 & [66.81, 67.66] & 66.80 & 81.31 & 43.20 & 0.0888 \\
\midrule
\addlinespace[3pt]
\multicolumn{7}{@{}l}{\textit{AnyJev on Qwen3-32B}} \\
Raw (one forward) & 67.03 & [66.60, 67.46] & 69.49 & 75.45 & \textbf{47.51} & 0.1982 \\
L0: rotations, prior off & 67.59 & [67.17, 68.02] & 70.40 & 75.96 & \textbf{47.51} & 0.1907 \\
L0: batch prior, $\alpha=0.75$ & 65.55 & [65.12, 65.99] & 68.16 & 72.99 & 47.48 & 0.1899 \\
Rotations + batch prior, $\alpha=1.0$ & 62.80 & [62.36, 63.24] & 64.95 & 70.10 & 45.86 & 0.2063 \\
Rotations + content-free, $\alpha=0.5$ & 63.51 & [63.07, 63.95] & 70.39 & 66.56 & 45.10 & 0.2388 \\
Rotations + content-free, $\alpha=0.75$ & 61.62 & [61.18, 62.07] & 70.06 & 62.39 & 44.31 & 0.2613 \\
Rotations + content-free, $\alpha=1.0$ & 59.24 & [58.79, 59.68] & 68.88 & 58.20 & 42.82 & 0.2854 \\
\bottomrule
\end{tabular*}
\endgroup

  \par\vspace{5pt}
  \begin{minipage}{\textwidth}
  \footnotesize
  \textit{Notes.} Every measured row contains 46\,320 decisions. Accuracy pools correct counts
  across shards. ECE uses 15 equal-mass bins within each shard; the reported value is their mean
  (four shards for AnyJev and Nimble, one for the other baselines). It is not pooled ECE.
  Wilson intervals treat decisions as independent and do not account for repeated states or
  questions; they are descriptive, not a test of between-system differences.
  The majority reference chooses its class from test labels within each question-ID/option-count
  group and is an oracle diagnostic, not a deployable baseline. Uniform random is its analytic
  expected accuracy. Neither reference has a reported confidence interval or ECE.
  \end{minipage}
\end{table}

\paragraph{Readout protocol.} Raw reads the listed order once. L0 with the prior off averages
every cyclic rotation in log space; ordered score levels retain their original order, so their
raw and rotation results coincide. The batch prior subtracts the question-wise mean option
log-score within each evaluation shard at the displayed strength, for groups of at least 20
items; it uses test inputs but no test labels. The content-free alternative subtracts scores
from three neutral inputs (empty text, \texttt{N/A}, and \texttt{[MASK]}). Both are applied after
rotation averaging. No temperature is fitted, and no row represents L1 or the supervised L2 head.

\paragraph{Systems and interpretation.} AnyJev keeps the released base-model weights frozen.
Nimble~\cite{nimble}, Laya~\cite{laya} and NanoJev~\cite{nanojev} use their published prediction
interfaces; Qwen3-Reranker-0.6B scores each option separately using its yes/no reranking template. 
Question labels are excluded from model inputs. The dataset stores each label inside the same object as its question text, so the one run made before that exclusion was in place, Laya, was repeated with it; the two runs agree on all 46 320 decisions. None of the four baselines was trained on this dataset: Nimble on its own synthetic set, Laya on \texttt{LocalLLaMA/typed-decisions}, NanoJev on recorded expert episodes from four game environments, and Qwen3-Reranker on retrieval data. Their results here measure transfer rather than performance on the workloads they were built for. 
Nimble uses a Qwen3.5-9B LoRA adapter, so it shares
AnyJev's base architecture but not its final weights or prompt. Its 70.57\% overall accuracy
exceeds the best AnyJev result of 68.01\%; the latter has lower ECE (0.0490 versus 0.0889).
The two systems are close on yes/no decisions (81.37\% for AnyJev versus 81.06\% for Nimble),
while Nimble leads on choices (73.60\% versus 67.69\%). All measured systems have their lowest
accuracy on ordinal scores. Every tested prior correction lowers overall accuracy relative to
rotation averaging alone on its respective base model, even where ECE improves.

\paragraph{The evaluated L2-mono variant.} This run fits an affine map to the model's own
full-depth states on 1\,500 training-split prompts, without gold labels. The first two thirds
fit the map; the remaining third selects ridge strength and the earliest tested block reaching
95\% agreement with the full-depth, single-forward decision. It selects block 29/32, with
95.6\% calibration agreement. Its 67.24\% test accuracy is 0.61 points below the raw single
forward and 0.77 points below the rotation average. The 9.4\% reduction in block count is not
a measured latency improvement, and the calibration agreement is not a test-set guarantee.

\paragraph{Scope.} These measurements use two AnyJev bases on one dataset configuration.
We use Transformers 4.55 for Qwen3-32B and Qwen3-Reranker-0.6B, and 5.17 for
Qwen3.5-9B, Nimble and NanoJev; Laya uses its own package. AnyJev prompts are capped at 2\,048 tokens, whereas the
Nimble interface supports up to 8\,192, so this is a comparison of evaluated systems rather than
an equal-context ablation. No cross-system latency or memory comparison was measured.

\end{document}